\documentclass[twoside,leqno,twocolumn]{article}

\usepackage[letterpaper]{geometry}

\usepackage{ltexpprt}

\usepackage{amsmath,amssymb,amsfonts}
\usepackage{graphicx}
\usepackage{textcomp}
\usepackage{xcolor}
\usepackage{booktabs} 
\usepackage{amsmath}
\usepackage{graphicx}
\usepackage{subfigure}
\usepackage{multirow}
\usepackage{amsfonts,bm}
\usepackage{braket}
\usepackage{mathtools}
\usepackage{enumerate}
\usepackage{enumitem}
\usepackage{empheq}
\usepackage{calc}
\usepackage{balance}
\usepackage{dsfont}
\usepackage{textcomp} 
\usepackage{caption}
\usepackage{color, colortbl}
\usepackage{url}

\begin{document}



\title{\Large Modeling Reservoir Release Using Pseudo-Prospective Learning and Physical Simulations to Predict Water Temperature}

\author{Xiaowei Jia$^{1}$, Shengyu Chen$^{1}$, Yiqun Xie$^{2}$, Haoyu Yang$^3$, Alison Appling$^4$, \\Samantha Oliver$^4$, Zhe Jiang$^5$\\
\small\baselineskip=9pt $^1$ University of Pittsburgh, 
\small\baselineskip=9pt $^2$ University of Maryland, 
\small\baselineskip=9pt $^3$ University of Minnesota,\\ 
\small\baselineskip=9pt $^4$ U.S. Geological Survey, 
\small\baselineskip=9pt $^5$ University of Florida\\
}

\date{}

\maketitle


\fancyfoot[R]{\scriptsize{Copyright \textcopyright\ 20XX by SIAM\\
Unauthorized reproduction of this article is prohibited}}





\begin{abstract} \small\baselineskip=9pt 
This paper proposes a new data-driven method for predicting water temperature in stream networks with reservoirs. The water flows released from reservoirs greatly affect the water temperature of downstream river segments. However, the information of released water flow is often not available for many reservoirs, which makes it difficult for data-driven models to capture the impact to downstream river segments. In this paper, we first build a state-aware graph model to represent the interactions amongst streams and reservoirs, and then propose a parallel learning structure to extract the reservoir release information and use it to improve the prediction. In particular, for reservoirs with no available release information, we mimic the water managers' release decision process through a pseudo-prospective learning method, which infers the release information from anticipated water temperature dynamics. For reservoirs with the release information, we leverage a physics-based model to simulate the water release temperature and transfer such information to guide the learning process for other reservoirs.  
The evaluation for the Delaware River Basin shows that the proposed method brings over 10\% accuracy improvement over  existing data-driven models for stream temperature prediction when the release data is not available for any reservoirs. The performance is further improved after we incorporate  
the release data and physical simulations for a subset of reservoirs.   
\end{abstract}

\section{Introduction}

Accurate predictions of water temperature in streams are critical for many decision making processes since water temperature is directly related to important aquatic outcomes, including the suitability of aquatic habitats and  greenhouse gas exchange~\cite{brett1971energetic}.  
The objective of this paper is to predict water temperature for all the river segments in a stream network at a daily scale.  
This is 
a challenging problem since water temperature in streams is affected by a combination of complex processes including weather (e.g., air temperature, solar radiation, precipitation), interactions between connected river segments in the stream network, and the process of water release from reservoirs~\cite{bogan2003stream}. In particular, the water flow released from reservoirs  can greatly impact water temperature for downstream river segments. For example, resource managers often release cold water from the bottom of the reservoirs to reduce downstream river water temperatures, which is needed for maintaining 
desired temperature regimes for aquatic life.

One intuitive approach for stream temperature prediction is to build individual models for each river segment combining the information of weather and its upstream river segments and reservoirs. However, this would be challenging given the resources necessary to collect sufficient water temperature observations for each river segment. 
Moreover, the explicit information of upstream streams and reservoirs, e.g., the amount of water flow advected to the downstream river segment, are often not available. 

\begin{figure} [!t]
\centering
\subfigure[]{ \label{fig:b}{}
\includegraphics[width=0.47\columnwidth]{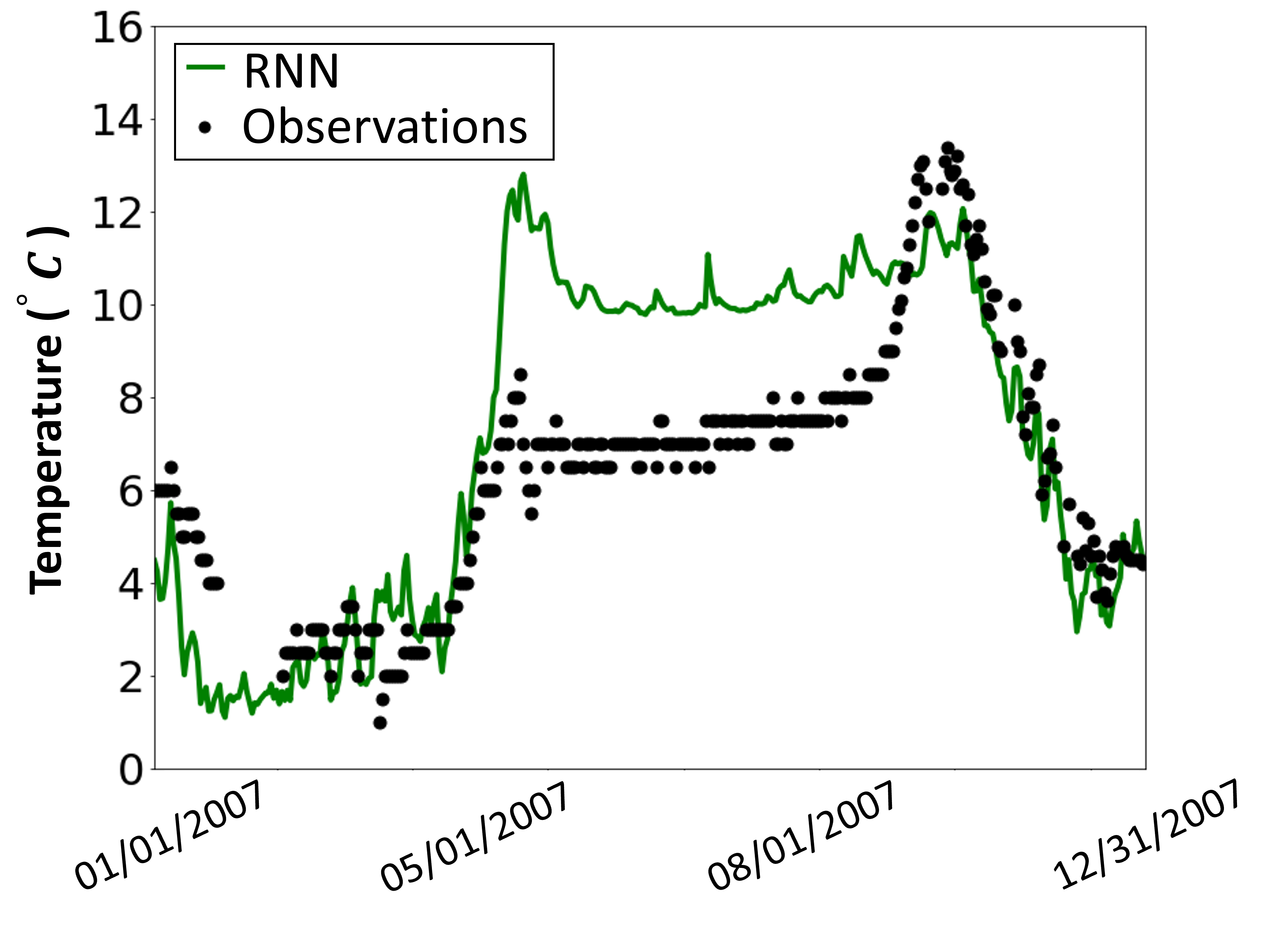}
}\hspace{-.25in}
\subfigure[]{ \label{fig:b}{}
\includegraphics[width=0.5\columnwidth]{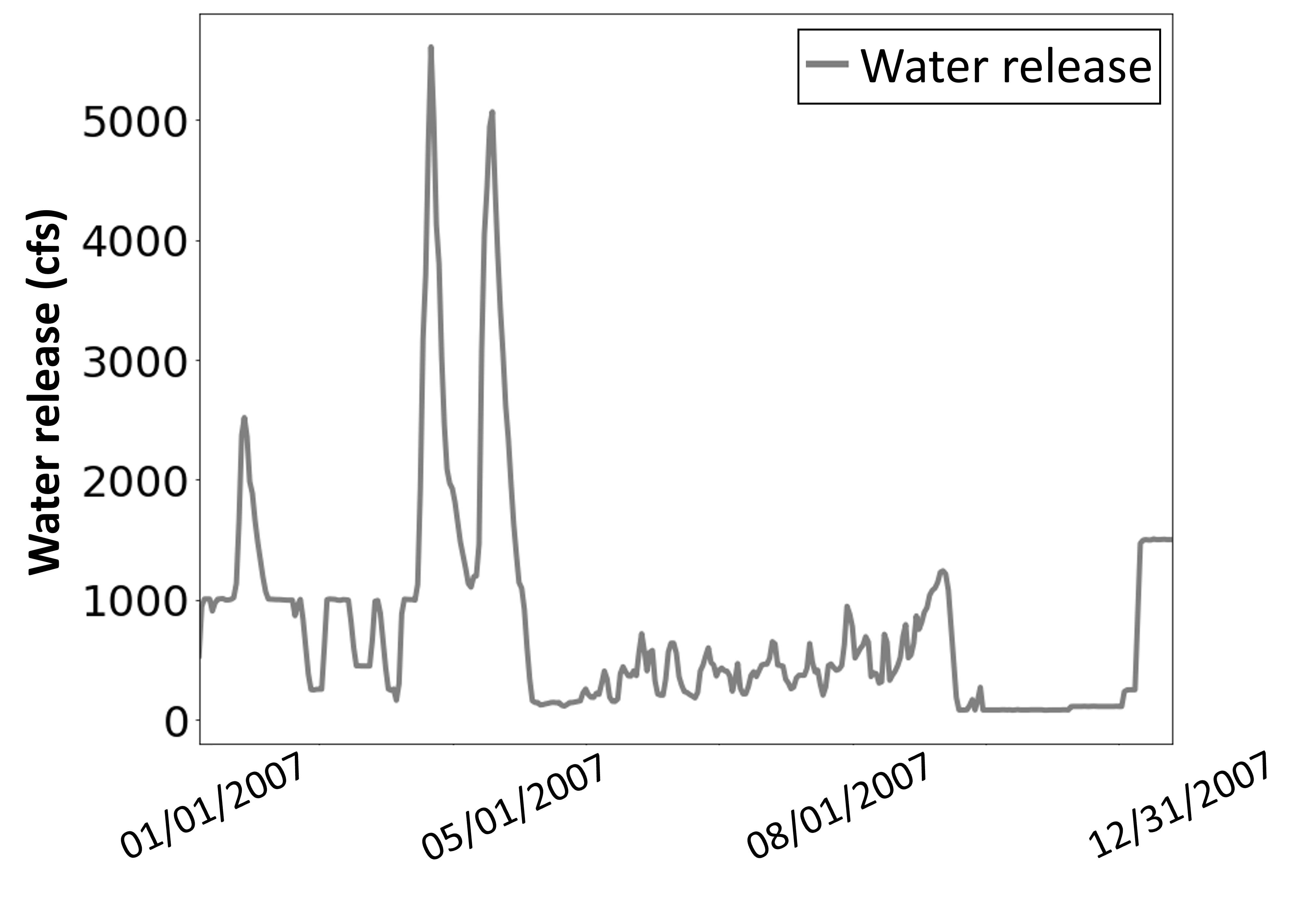}
}\vspace{-.2in}
\caption{(a) RNN predictions on a river segment in 2007. (b) The amount of water released from a reservoir upstream from the river segment in (a). }
\label{fig:exp}
\end{figure}

Prior work has used global data-driven models~\cite{moshe2020hydronets,jia2021physics} to simulate water temperature dynamics in the entire stream network given variations in climate drivers (e.g., solar radiation, precipitation, and air temperature), catchment characteristics, and the influence of the stream network topology. 
However, these approaches are not designed for capturing the impact of reservoirs, which 
brings stochasticity  to 
the relationship between input climate drivers and observed water temperature, and makes it difficult 
to capture such relationship. Fig.~\ref{fig:exp} shows the predictions made by a global recurrent neural network model (RNN) with a Long-Short Term Memory (LSTM) cell. The RNN model significantly over-estimates the water temperature when cold water is released from an upstream reservoir at the beginning of the summer. 

The objective of this paper is to develop a new data-driven method to model  the impact of reservoirs and improve the prediction on all the river segments in stream networks. 
There are two major challenges faced by existing machine learning algorithms when applied to this problem. \textit{Challenge 1:}   
The water temperature in a river segment is impacted by upstream river segments and reservoirs. Similarly, the water temperature profile of a reservoir can also be affected by the water flow from its upstream rivers. The patterns of these interactions are different since reservoirs and streams have very different physical properties. In particular, reservoirs are man-made lakes formed upstream from  dams, and they commonly have stratified layers with different temperature  while streams are  shallower and usually assumed to be well-mixed. The data-driven model needs to explicitly represent such differences when representing their interactions. 
\textit{Challenge 2:} Reservoir release data, e.g., how much water is released on each date and which depth layer is the water released from,  is often not available for many reservoirs due to privacy concerns.  
This poses a challenge for existing data-driven approaches to model the impact of reservoirs on downstream river segments. 

In this work, we build a State-Aware Graph (SAG) model, which maintains different state variables for river segments and reservoirs and uses a graph structure to represent their interactions. We also propose a pseudo-prospective (PP)  approach to embed the release information  when the reservoir release data is not available. 
In practice, water managers often release cold water from the bottom of a reservoir if they anticipate water temperature for downstream segments going above a threshold~\cite{ravindranath2016environmental_backup}. 
The idea of the PP approach is to mimic this process by 
referring to the future water temperature assuming no water flow is released from reservoirs. Since  the future water temperature is not accessible in practice, we build a forecasting model to produce anticipated water temperature, i.e., pseudo-prospective water temperature.

The PP approach also has its own limit due to the uncertainty from the forecasting model. Moreover, the PP approach does not consider other potential factors that affect the water release management, such as the water supply to surrounding cities and expected position of the salt front. 
To further improve the prediction, we also leverage the release data available from certain ``transparent'' reservoirs and transfer the knowledge of underlying physical processes from these ``transparent'' reservoirs to other reservoirs. In particular, we run a physics-based General Lake Model~\cite{hipsey2019general} on ``transparent'' reservoirs to simulate their water temperature dynamics for different depth layers at a daily scale. Then we combine the simulated temperature profiles and the release data (i.e., how much water is released from each depth layer on each date) to estimate the temperature of the released water. Such information directly reflects the impact of reservoir release to the downstream river segments. We  use a simulation-based embedding (SE) approach to include such information in the SAG model and also build a parallel structure to transfer the information from ``transparent" reservoirs to other reservoirs for which the PP approach is used.

We evaluate the proposed method for the Delaware River Basin. 
The results demonstrate that the model performs well in three scenarios  (1) when no reservoir release information is provided for all the reservoirs, (2) when a subset of reservoirs do not have the release data, and (3) when the model is generalized to a new stream network that is spatially disjoint to the training region and has no reservoir release data. 


\section{Related Work}


Graph neural networks have been applied to multiple scientific problems and shown improved predictive performance~\cite{qi2019hybrid,xie2018crystal,zhu2020understanding}. 
These advances have enhanced the capability to model interacting processes in complex physical systems, 
which commonly requires substantial efforts in calibration in traditional physics-based modeling approaches. 
Graph neural networks have also shown 
potential for the modeling of water temperature and streamflow in river networks~\cite{jia2021physics,moshe2020hydronets}. 
Despite the accuracy improvement brought by these methods, they are mostly evaluated in stream regions without reservoirs. The performance of these methods can be impacted when reservoirs are present in the stream networks but unaccounted for in the graph network.   

The graph model used in this paper is inspired by the heterogeneous graph, which is commonly used to represent multiple types of connections amongst multiple types of nodes~\cite{shi2016survey}. Neural network models have also been developed to represent such a graph structure and discover knowledge from heterogeneous data~\cite{zhang2019heterogeneous,wang2019heterogeneous,zhu2020hgcn}. Our previous paper~\cite{chen2021heter} also used heterogeneous graphs to represent the complex stream networks with both river segments and reservoirs.  
Compared to convolutional neural networks (CNNs), the graph-based model is more flexible in representing spatial dependencies amongst irregularly distributed locations, which are common in environmental applications. Moreover, the graph-based model can be used as a building block  and  combined with other models, e.g., Long-Short Term Memory (LSTM), in neural networks to capture other types of data dependencies.  
Despite its capabilities, little is known on how graph-based models can be used to represent multiple complex interactions amongst different types of processes in scientific problems. 
The nature of scientific studies requires adaptation of these neural network models based on scientific knowledge to better represent the influence amongst processes. 

Prior works have shown the potential for combining physical simulations with machine learning models. For example, simulated data can be used to pre-train deep learning models~\cite{jia2019physics,sultan2018transferable,ham2019deep,read2019process} and add supervision to intermediate hidden variables~\cite{khandelwal2020physics,jia2021physics}. These studies have shown improved model accuracy and generalizability using limited observed samples.

\section{Problem definition}
We consider $N$ river segments and $M$ reservoirs in a stream network. For each river segment $i$, we are provided with input features over multiple daily time steps $\textbf{X}_i=\{\textbf{x}_{i}^{1}, \textbf{x}_{i}^2, ..., \textbf{x}_{i}^T\}$.  Here input features $\textbf{x}_{i}^{t}$ form a $D_x$-dimensional vector, which includes climate drivers and geometric parameters of the segment (more details can be found in Section~\ref{sec:dataset}). For each reservoir $k$, we are  provided with its static $D_m$-dimensional meta-features $\textbf{l}_k$ such as the height and width of the dam. We also have observed temperature $\textbf{Y}=\{{y}_i^t\}$ for certain segments and on certain dates.   Our objective is to predict water temperature over multiple river segments in the stream network at a daily scale by leveraging the spatial and temporal contextual information.  

We use a graph $\mathcal{G} = \{\mathcal{V},\mathcal{E},\textbf{A}\}$ to represent dependencies amongst river segments and reservoirs. 
Here the node set $\mathcal{V}=\{\mathcal{V}_s,\mathcal{V}_r\}$ contains the set of river segments $\mathcal{V}_s$ and reservoirs $\mathcal{V}_r$. The edge set $\mathcal{E}=\{\mathcal{E}_{ss},\mathcal{E}_{sr},\mathcal{E}_{rs}\}$ contains three types of edges among river segments and reservoirs. Specifically, $\mathcal{E}_{ss}$ represents the edges between pairs of segments $(i,j)$ where the segment $i$ is anywhere upstream from the segment $j$, $\mathcal{E}_{sr}$ represents the edges between river segments and their downstream reservoirs, and $\mathcal{E}_{rs}$ represents the edges between reservoirs and their downstream river segments. 
The matrix $\textbf{A}\in \mathbb{R}^{(N+M)\times (N+M)}$ represents the adjacency level between each pair of river segments or between river segments and reservoirs in the graph. Specifically, $\textbf{A}_{ij}=0$ means there is no connection from node $i$ to node $j$ and a higher value of $\textbf{A}_{ij}$ indicates that the node $i$ is closer to node $j$ in terms of the stream distance. 
More details of how we generate the adjacency matrix are discussed in Section~\ref{sec:eva_details}.

\section{Method}
The methods proposed in this paper aim to tackle three sub-tasks: (i) how to represent streams and reservoirs using neural networks, (ii) how to model the impact of reservoirs when their release data are not available, and (iii) how to further leverage the simulated data produced by physics-based models. 
In Section 4.1, we first introduce the state-aware graph (SAG) model to represent how river segments and reservoirs evolve and interact with each other. Then in Section 4.2, we discuss the pseudo-prospective method to infer the reservoir release information when it is not available. Finally, in Section 4.3, we describe how to leverage the release information available at certain reservoirs and physical simulations to further improve the model performance. 
The model code and outputs are available from~\cite{jia2022datarelease}.

\subsection{State-Aware Graph (SAG) Model}

\begin{figure} [!t]
\centering
\subfigure[]{ \label{fig:b}{}
\includegraphics[width=0.25\columnwidth]{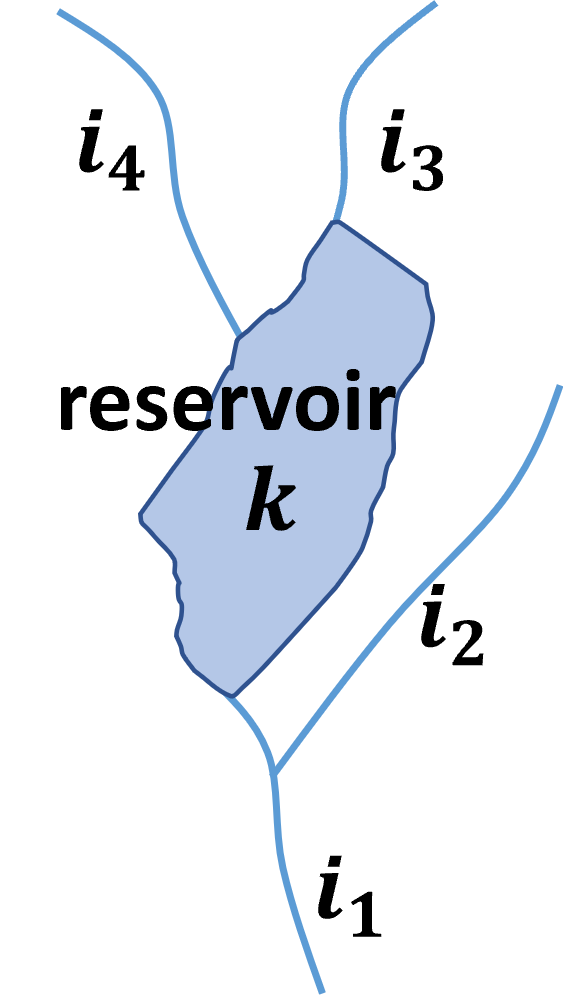}
}
\subfigure[]{ \label{fig:b}{}
\includegraphics[width=0.6\columnwidth]{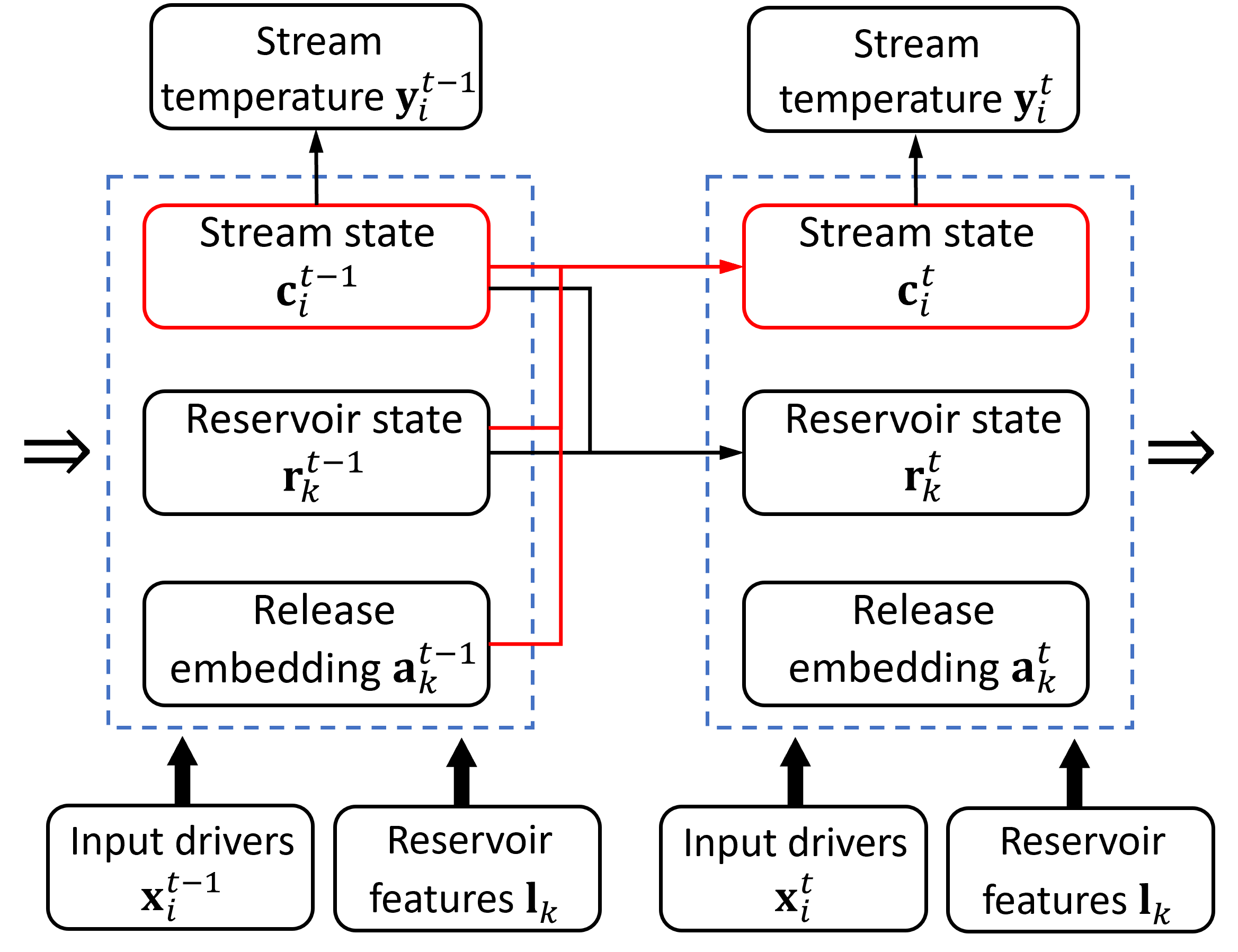}
}\vspace{-.2in}
\caption{(a) An example stream network with a reservoir. (b) The structure of SAG model. Each recurrent unit maintains the stream states $\textbf{c}$ and the reservoir states $\textbf{r}$. 
The figure shows the update mechanism of stream state $\textbf{c}$ and reservoir state $\textbf{r}$ between two units. The arrow indicates the edge in the computation graph (Eqs.~\ref{eq:res-state}-\ref{eq:predict}). The computation of the release embedding $\textbf{a}$ is shown in Fig.~\ref{fig:PP}. }
\label{fig:flow}
\end{figure}

Streams and reservoirs have different temperature patterns while also being affected by each other, i.e., stream water flowing into a reservoir affects the reservoir's temperature, and water release from reservoirs also affects the temperature of downstream river segments. Hence, the machine learning model needs to memorize the state of reservoirs and streams over time and capture their interactions.  
The intuition of the SAG model is to use two sets of state variables (stream states $\{\textbf{c}_i\}$ and reservoir states $\{\textbf{r}_k\}$, both of dimension $D_h$) to capture how streams and reservoirs evolve and interact with each other (Fig.~\ref{fig:flow}). The state variable for each river segment or reservoir is a multi-dimensional vector that encodes the influence of weather and the spatio-temporal context. 
In the following, we describe how to update state variables over time.





\noindent\underline{\textit{State of reservoirs:}} Since water flows from upstream river segments can change the temperature of reservoirs, we update the reservoir state $\textbf{r}_k^{t}$ for each reservoir $k$ at time $t$  by incorporating the influence from  its upstream river segments at  the previous time step $t-1$. The change of reservoir temperature given such influence also  depends on the characteristics of the reservoir (e.g., the geometry of reservoirs). Hence, for each reservoir, we combine the state variables $\textbf{c}_i$ of its upstream river segments (represented as $\mathcal{S}(k)$) and use the static features $\textbf{l}_k$ to filter the influence from these river segments  before updating the reservoir state, as: 
\begin{equation}
\small
\textbf{r}_{k}^{t} = \text{tanh}(\textbf{W}_{r} \textbf{r}_{k}^{t-1}+ f_1(\textbf{l}_k)\odot \sum_{i\in \mathcal{S}(k)}\textbf{A}_{ik}\textbf{c}_i^{t-1} +\textbf{b}_{r}),
\label{eq:res-state}
\vspace{-.05in}
\end{equation}
where 
$\textbf{W}\in \mathbb{R}^{D_h\times D_h}$ and $\textbf{b}\in \mathbb{R}^{D_h}$ are model parameters, 
$\odot$ represents the element-wise product, the function $f_1(\cdot)$ transforms the static meta-features of the reservoir to the same dimension with hidden variables with each output variable in the range of [0,1]. 
We implement $f_1(\cdot)$ using fully connected layers and the sigmoid activation function. Here the influence of each upstream river segment is also weighted by its adjacency level to the reservoir.

\noindent\underline{\textit{State of river segments:}} 
For each river segment $i$, its water temperature at time $t$ is affected by (1) the stream state at the previous time, (2) the weather at the current time, (3) the water advected from upstream reservoirs, and (4) the water advected from upstream river segments. 
Similar to LSTM~\cite{hochreiter1997long}, we use multiple gating variables to filter the information from different sources and then combine the filtered information to update the stream state $\textbf{c}_i^t$. 
This is analogous to the evolution of a dynamical system, in which the state of streams changes over time in response to influences from different sources (e.g., solar radiation, advected water, etc.) filtered by specific physical conditions. This process is shown as: 
\begin{equation}
\small
\begin{aligned}
\textbf{c}_i^t &= \text{tanh}(\textbf{gf}_i^t\odot \textbf{c}_i^{t-1}+\textbf{gi}_i^t\odot\bar{\textbf{c}}_i^t+\textbf{gr}_i^t\odot \textbf{p}^{t-1}_i+\textbf{gs}_i^t\odot \textbf{q}^{t-1}_i),
\end{aligned}
\label{eq:stream-state}
\vspace{-.05in}
\end{equation}
where $\textbf{gf}_i^t$, $\textbf{gi}_i^t$, $\textbf{gr}_i^t$, $\textbf{gs}_i^t$ represent the gating variables used to filter the information from historical stream states, the current weather input, upstream reservoirs, and upstream river segments, respectively.  The candidate state  $\bar{\textbf{c}}_i^t$  encodes the information of river segment $i$ at the current time $t$, $\textbf{p}_i^{t-1}$ and  $\textbf{q}_i^{t-1}$ are the latent  variables (referred to as transferred variables) that embed the effect from upstream reservoirs and river segments, respectively. We use the transferred variables from the previous time step to account for the water travel time. We now describe how to compute these variables.


We first follow the same process in LSTM~\cite{hochreiter1997long} to compute the candidate state $\bar{\textbf{c}}_i^t$  by combining climate drivers at the current time step ${\textbf{x}}_i^t$ and the hidden representation at previous time step ${\textbf{h}}_i^{t-1}$ (computed from ${\textbf{c}}_i^{t-1}$ by Eq.~\ref{eq:hidden}), as follows: 
\vspace{-.03in}
\begin{equation}
\small
\begin{aligned}
\bar{\textbf{c}}_i^t &= \text{tanh}(\textbf{W}_c^h \textbf{h}_i^{t-1} + \textbf{U}_c^x \textbf{x}_i^t+\textbf{b}_c), 
\end{aligned}
\end{equation}
where $\textbf{U}\in \mathbb{R}^{D_h\times D_x}$ denotes model parameters to transform input data.

For a river segment, the impact it receives from a reservoir depends on the reservoir state and its characteristics (e.g., reservoir depth), as well as the water release information, e.g., the volume and temperature of the water flow  released from the reservoir. 
In particular, the water release information is critical for modeling the impact since the water managers can adjust the amount of released water to control the change of downstream temperature.   
We create a  release embedding $\textbf{a}_i^k\in \mathbb{R}^{D_h}$ to encode the water release information. The computation of the release embedding $\textbf{a}_i^k$ is challenging due to the missing release data for many reservoirs, which will be addressed in Sections~\ref{sec:pp} and~\ref{sec:se}.
If release embeddings from each reservoir are available, we compute the transferred  variables $\textbf{p}_i^{t-1}$ for a river segment $i$ combining the information from its upstream reservoirs (represented as $\mathcal{M}(i)$) as:
\begin{equation}
\small
\textbf{p}_{i}^{t-1} = \text{tanh}(\textbf{W}_p\!\!\sum_{k \in \mathcal{M}(i)}\!\!\textbf{A}_{ki}f_2(\textbf{l}_k)\odot(\textbf{W}_p^{r} \textbf{r}^{t-1}_{k}+\textbf{a}^{t-1}_k)+\textbf{b}_p). 
\label{eq:trans_p}
\vspace{-.05in}
\end{equation}
where 
$f_2(\cdot)$ is also used to convert static features of the reservoir to the filtering variables and is implemented using fully connected layers. 



For each river segment $i$, we also use transferred variables $\textbf{q}_i^{t-1}$ to capture the impact from its upstream river segments   
(represented as $\mathcal{N}(i)$)  as follows:
\begin{equation}
\small
\textbf{q}_{i}^{t-1} = \text{tanh}(\textbf{W}_q\sum_{j \in \mathcal{N}(i)}\textbf{A}_{ji}\textbf{h}_{j}^{t-1}+\textbf{b}_q). 
\end{equation}

Then we generate four sets of  gating variables 
using the sigmoid function $\sigma(\cdot)$ as follows: 
\begin{equation}
\footnotesize
\begin{aligned}
\textbf{gf}_i^t &= \sigma(\textbf{W}_f^h \textbf{h}_i^{t-1} + \textbf{U}_f^x \textbf{x}_i^t+\textbf{b}_f),\\
\textbf{gi}_i^t &= \sigma(\textbf{W}_g^h \textbf{h}_i^{t-1} + \textbf{U}_g^x \textbf{x}_i^t+\textbf{b}_g),\\
\textbf{gr}_i^t &= \sigma(\textbf{W}_r^p \textbf{p}^{t-1}_{i} + \textbf{U}_r^x \textbf{x}_i^t+\textbf{b}_r),\\
\textbf{gs}_i^t &= \sigma(\textbf{W}_s^q \textbf{q}^{t-1}_{i} + \textbf{U}_s^x \textbf{x}_i^t+\textbf{b}_s).
\end{aligned}
\end{equation}



After obtaining the stream state $\textbf{c}_i^t$ (Eq.~\ref{eq:stream-state}), we generate the output gating variables $\textbf{o}^t_i$ and use them to filter the model state to generate the hidden representation $\textbf{h}^t$, as follows:
\begin{equation}
\small
\begin{aligned}
\textbf{o}_i^t &= \sigma(\textbf{W}_o^h \textbf{h}_i^{t-1} + \textbf{U}_o^x \textbf{x}_i^t+\textbf{b}_o),\\
\textbf{h}_i^t &= \textbf{o}_i^t\odot \text{tanh}(\textbf{c}_i^t). 
\end{aligned}
\label{eq:hidden}
\vspace{-.05in}
\end{equation}


Finally, we generate predicted target variables $\hat{\textbf{y}}_i^t$ from the hidden representation, as follows:
\begin{equation}
\small
\hat{{y}}_i^t = \textbf{V}\textbf{h}_i^t+c,
\label{eq:predict}
\vspace{-.05in}
\end{equation}
where $\textbf{V}$ and $c$ are model parameters. 

%
The SAG model is trained to minimize the mean squared loss between observed temperature $\textbf{Y}=\{\textbf{y}_{i}^{t}\}$ and predicted values. The loss is only measured at certain time steps and locations for which observations are available. 


\subsection{Pseudo-prospective reservoir embedding}
\label{sec:pp}

\begin{figure} [!t]
\centering
\includegraphics[width=0.9\columnwidth]{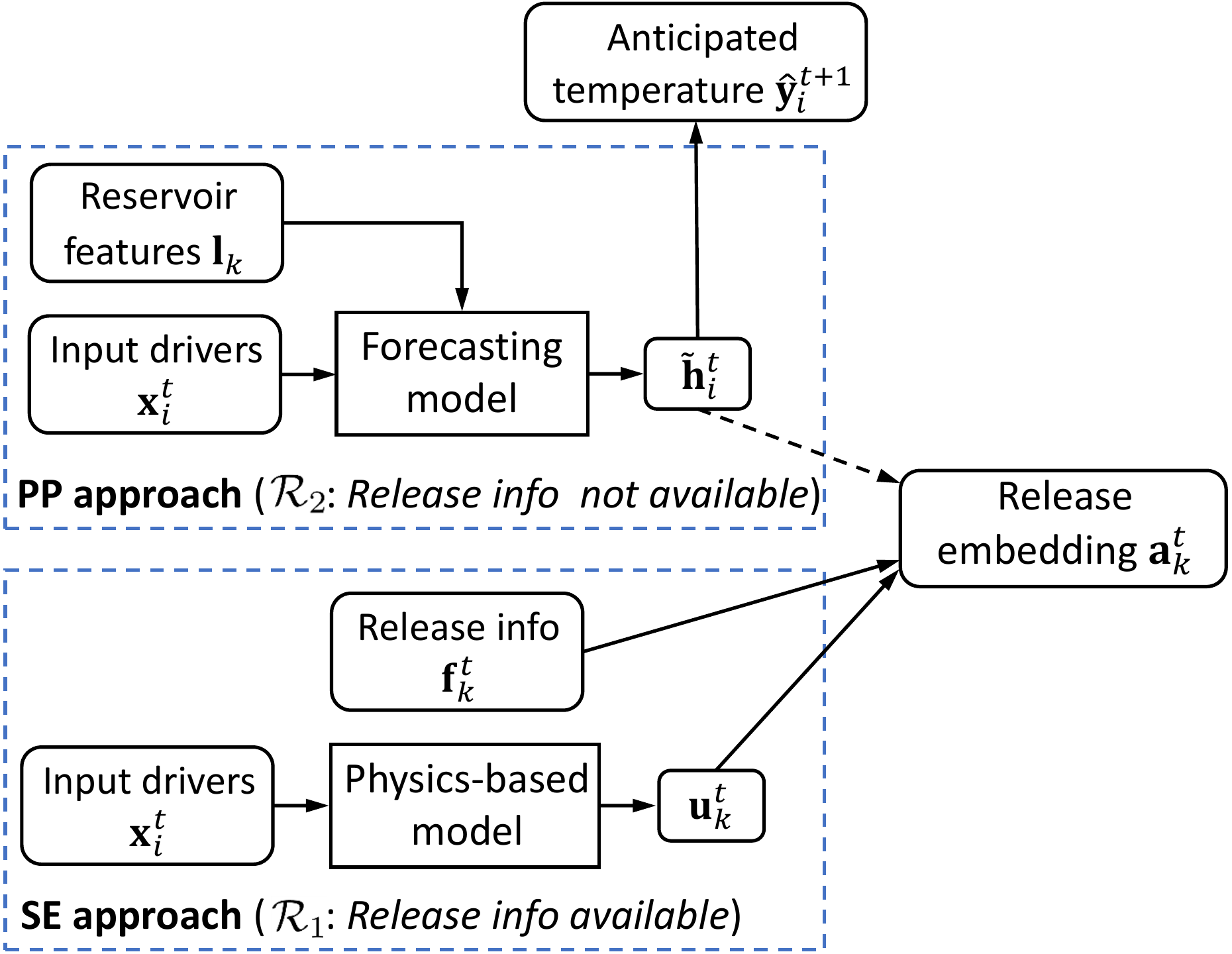}
\vspace{-.15in}
\caption{The parallel structure to compute the release embeddings. 
The SE approach is used for reservoirs with the release information ($\mathcal{R}_1$) while the PP approach is used for 
reservoirs without the release information ($\mathcal{R}_2$). 
The dashed line between the PP approach and the release embeddings implies that we use a separate forecasting model to learn the hidden representation $\tilde{\textbf{h}}$ of the anticipated water temperature.}
\label{fig:PP}
\end{figure}

One major challenge in building the SAG model is that we do not have access to the release data for many reservoirs. Since the water release from reservoirs often has a much lower temperature, the prediction can be biased if the model does not consider the water flow from reservoirs. Hence, we aim to design a new mechanism to compensate for the missing reservoir release information. 

One major objective for reservoir release is to maintain the desired water temperature for the suitability of aquatic habitat~\cite{ravindranath2016environmental}. 
Managers often make decisions to release water from a reservoir based on the anticipated water temperature in the future (by a separate model)~\cite{ravindranath2016environmental}. For example, for Cannonsville Reservoir, the water managers will release cold water from the lower depth of the reservoir when they anticipate the next day's water temperature for downstream rivers will be above 75$^\circ$~\cite{flexiblemanagement}. 

We create a PP learning approach to mimic such reservoir release processes (the upper block of Fig.~\ref{fig:PP}). The idea of PP learning is inspired by the prior work~\cite{yang2022strategies}, which aims to improve the learning task at the current time by leveraging anticipated future information that is unavailable in real scenarios. 
In our problem, we create the release embedding $\textbf{a}_k^{t}$ using the information of anticipated water temperature for river segments that are downstream from the reservoir $k$. 

We first create a separate stream temperature forecasting model, which uses the input features at the current time step $t$ to predict the water temperature at the next time step $t+1$. In particular,  we use input-output pairs $\{(\textbf{x}^t, \textbf{y}^{t+1})\}$ from the training data for training this forecasting model. To ensure the forecasting model provides unbiased anticipations for reservoir release decisions, we do not use training samples from river segments downstream from reservoirs. 
This model uses the same structure as the SAG model except that (1) it outputs $\textbf{y}^{t+1}$ for input $\textbf{x}^t$, and (2) it does not consider the release embedding, i.e., the transferred variables $\textbf{p}_i^{t-1}$ (originally computed by Eq.~\ref{eq:trans_p}) become:  
\begin{equation}
\small
\textbf{p}_{i}^{t-1} = \text{tanh}(\textbf{W}_p\!\!\sum_{k \in \mathcal{M}(i)}\!\!\textbf{A}_{ki}f_2(\textbf{l}_k)\odot\textbf{W}_p^{r} \textbf{r}^{t-1}_{k}+\textbf{b}_p).\label{eq:trans_p_pp}
\end{equation}


This forecasting model connects the information at the current time to the anticipated information in the future that could inform water mangers' release decisions. 
We apply the forecasting model to each time step and obtain the extracted hidden representation $\tilde{\textbf{h}}_i^{t}$ from the forecasting model (Eq.~\ref{eq:hidden}), which embeds the information about the anticipated water temperature   for each river segment $i$. We then combine $\tilde{\textbf{h}}_i^{t}$ from all the segments that are anywhere downstream from a reservoir $k$ (represented as $\mathcal{S}_{dn}(k)$) to generate its release embedding $\textbf{a}_k^{t}$, as: 
\vspace{-.05in}
\begin{equation}
\small
    \textbf{a}_k^{t} = 
    \sum_{i\in \mathcal{S}_{dn}(k)}\textbf{A}_{ki}\textbf{W}_{pp} \tilde{\textbf{h}}_i^{t} + \textbf{b}_{pp}.
    \label{eq:pp}
    \vspace{-.05in}
\end{equation}


\subsection{Leveraging physical simulations and release data}
\label{sec:se}


The PP embedding still has limits in that the forecasting model is not fully accurate and also may not be consistent with the decision making process followed by water managers. 
We  leverage the reservoir release data that are available for certain reservoirs to further improve the prediction. The idea is to combine the release information and the knowledge of underlying physical processes  to better capture the impact of these reservoirs on their downstream river segments and also transfer the learned patterns to other reservoirs and river segments.

We first introduce a new release embedding by using the available release data and physical simulations. Then we create a parallel learning structure to 
transfer the knowledge learned from available release data and physical simulations to the PP embeddings on other reservoirs with no release information.    

\noindent\underline{\textit{Incorporating reservoir release and physical simulations:}}  
Here we introduce an SE approach to compute a new release embedding $\textbf{a}_k^t$ for reservoirs with the release information. 
In particular, we consider a subset of reservoirs for which we have the information of how much water (in cubic feet per second) is released from each depth layer $d\in \{1,2,...,L\}$  on each day $t$. For each reservoir $k$, we use $\textbf{f}^t_k$ to represent its release information on each date $t$, and it contains the release volume at multiple depth layers  $\textbf{f}^t_k = \{{f}^t_{k,1},{f}^t_{k,2},...,{f}^t_{k,L}\}$. 
To represent the impact of reservoir release to a downstream segment, we need to consider  both the amount of water flow from the reservoir 
and the temperature of the released water. However, the temperature of released water is driven by complex processes (e.g., vertical mixing, and the warming or cooling of water via energy lost or gained from fluxes such as solar radiation and evaporation) and also cannot be easily measured in practice. Hence, we will run a physics-based General Lake Model~\cite{hipsey2019general}  built based on  these underlying processes to simulate water temperature  at $D$ different depths of the reservoir $\{{m}^t_{k,1},{m}^t_{k,2},...,{m}^t_{k,L}\}$. Then we combine such simulations with the release flow information to compute a flow-average temperature as:
\begin{equation}
\small
    {u}^t_k = \frac{\sum_d {f}^t_{k,d} {{m}}^t_{k,d}}{\sum_d {f}^t_{k,d}}
    \label{eq:u}
\end{equation}

Combining the simulated flow-average temperature and the flow of the reservoir release, we generate the release embedding, as follows:
\begin{equation}
\small
    \textbf{a}_k^{t} = \textbf{Z} [\textbf{f}^t_{k},{u}^t_k] +\textbf{b}_{se},
    \label{eq:se}
\end{equation}
where $\textbf{Z}\in \mathbb{R}^{D_h\times (L+1)}$ is model parameters.

\noindent\underline{\textit{Transfer the knowledge to other segments:}} Compared to the release embedding generated through the PP approach (Eq.~\ref{eq:pp}), The SE embedding (Eq.~\ref{eq:se}) directly embeds the available information of the flow and temperature for released water, and thus  captures  the characteristics of reservoir release that affect downstream segments. %
However, it can be computed only for a subset of reservoirs  with the reservoir release information ($\mathcal{R}_1$). 
For reservoirs without the release information ($\mathcal{R}_2$), we need to use the PP embedding, and we transfer the knowledge learned from $\mathcal{R}_1$ to guide the PP embedding process. As shown in  Fig.~\ref{fig:PP}, we 
use PP and SE in parallel to generate release embeddings $\textbf{a}_k^t$ for reservoirs in $\mathcal{R}_1$ and $\mathcal{R}_2$, respectively. The generated release embeddings are then passed to shared layers to update stream states and make predictions (Fig.~\ref{fig:flow}). In this way, the information of reservoir release and the physical simulations used in the SE process can also regularize the PP embeddings as they need to be consistent with the outputs produced by the SE approach in how they impact stream states.


\begin{figure} [!t]
\centering
\includegraphics[width=0.9\columnwidth]{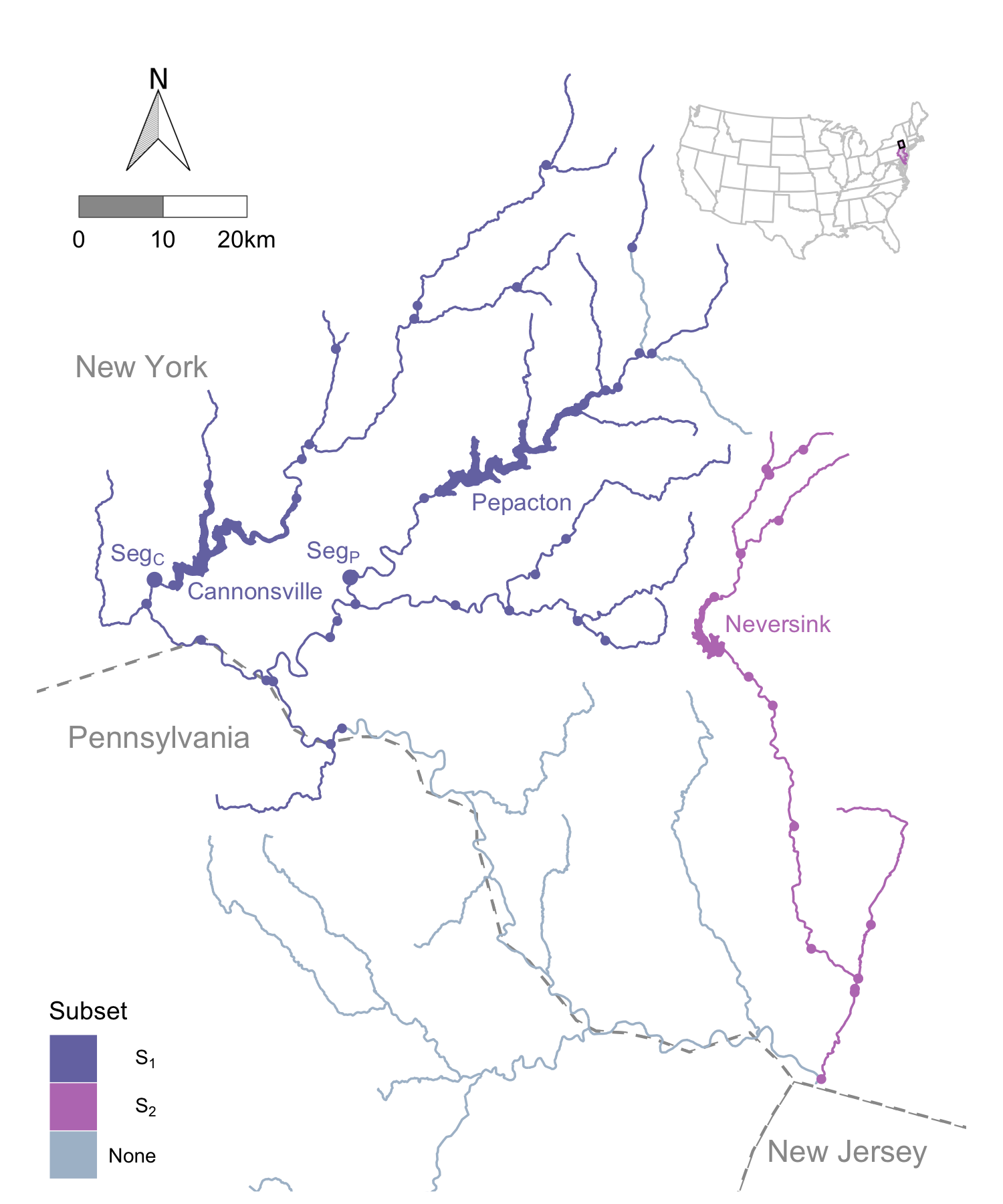}
\vspace{-.15in}
\caption{The river-reservoir network being modeled. Purple and pink indicate reaches are within study subsets $S_1$ and $S_2$, respectively, and gray reaches show the river network context for those subsets. Stream segment endpoints are marked with circles, and Seg$_C$, Seg$_P$, and the Cannonsville, Pepacton, and Neversink reservoirs have text labels. Dashed lines are state borders; water flows toward New Jersey. The inset shows the full Delaware River Basin in pink and the area of the main map in black.}
\label{fig:map}
\end{figure}

\begin{figure*} [!t]
\centering
\includegraphics[width=1.75\columnwidth]{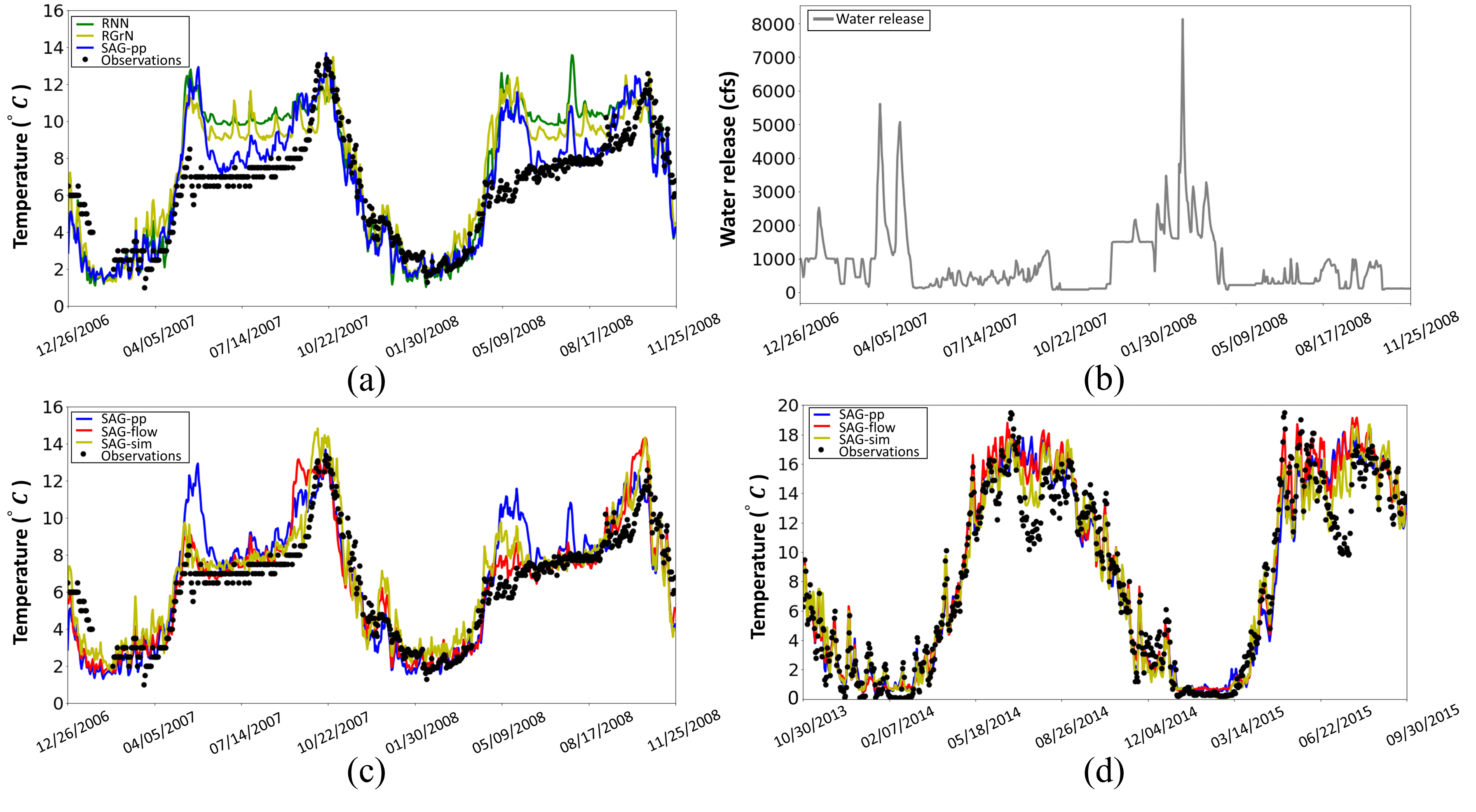}
\vspace{-.2in}
\caption{ (a) The predictions of RNN, RGrN, and SAG-pp on Seg$_C$. (b) The amount of reservoir release from the Cannonsville Reservoir (summed over all the depth layers). (c)-(d) The predictions of multiple SAG variants on Seg$_C$ (c) and Seg$_P$ (d). }
\label{fig:prd}
\vspace{-.2in}
\end{figure*}

\section{Experimental Results}
\label{sec:exp_res}

\subsection{Dataset}
\label{sec:dataset}
We evaluate the proposed method for predicting stream temperature using real-world data collected from the Delaware River Basin, which is an ecologically diverse region 
and a watershed along the east coast of the United States that provides drinking water to over 15 million people~\cite{williamson2015summary}.
The dataset used in our evaluation is from the U.S. Geological Survey's National Water Information System~\cite{us2016national} and the Water Quality Portal~\cite{read2017water}. 
Observations at a specific latitude and longitude were matched to river segments that vary in length from 48 to 23,120 meters. The river segments were defined by the geospatial fabric used for the National Hydrologic Model as described by Regan et al.~\cite{regan2018description}, and the river segments are split up to have roughly a one day water travel time. 
 

We study two spatially disjointed subsets of the Delaware River Basin (as shown in Fig.~\ref{fig:map}): The first subset $S_1$ includes 56 river segments flowing toward Lordville, NY, and the second subset $S_2$ includes 18 river segments flowing toward Sullivan County, NY. 
We select these subsets since we have sufficient observations collected in these areas.  
In particular, we use input features at the daily scale from Jan 01, 1980, to June 22, 2020 (14,784 dates). The input features have 10 dimensions which include daily average precipitation, daily average air temperature, date of the year, solar radiation, shade fraction, potential evapotranspiration and the geometric features of each segment (e.g., elevation, length, slope and width). Air temperature, precipitation, and solar radiation values were derived from the gridMET gridded meteorological dataset~\cite{gridMET}. Other input features (e.g., shade fraction, potential evapotranspiration) are difficult to  measure frequently, and we use values internally calculated by the physics-based PRMS-SNTemp model~\cite{theurer1984instream}. The subset  $S_1$ covers the Cannonsville and Pepacton Reservoirs. The release data includes how much water is released from specific depth layers at daily scale~\cite{jia2022datarelease}. In $S_1$, water temperature observations were available for 29 segments but the temperature was observed only on certain dates. The number of temperature observations  available 
for the 29 observed segments ranges from 1 to 13,000 with a total of 76,163  
observations across all dates and segments~\cite{jia2022datarelease}. The subset $S_2$ covers the Neversink reservoir. Water temperature observations were available for 16 segments in $S_2$, and the number of observations available ranges from 1 to 9,694 with a total of 21,846. 
For all the reservoirs, we also have meta-features of these reservoirs, including dam height, dam length, depth, elevation, and area of catchment~\cite{williamson2015summary}.


\subsection{Evaluation details}
\label{sec:eva_details}
We have released our dataset and implementation~\cite{jia2022datarelease}. 
The model is optimized with the ADAM optimizer~\cite{kingma2014adam} with the initial learning rate of 0.002. All the hidden variables and gating variables in SAG have 20 dimensions. 
We generate the adjacency matrix $\textbf{A}$ based on the stream distance between each pair of nodes. When measuring the distance between a pair of river segments $i$ and $j$, we use the stream distance $\text{dist}(i,j)$ between their outlets. 
We  standardize the stream distance and then compute the adjacency level as $\textbf{A}_{ij}=1/(1+\text{exp}(\text{dist}(i,j)))$ for each edge $(i,j)$.

We conduct experiments to answer three questions: \textit{Q1:  Can the proposed SAG model with the PP approach outperform existing methods when the reservoir release data is not accessible for any reservoirs?} We test the performance of the SAG model with the PP embedding approach (SAG-pp) in $S_1$, and compare it with multiple baselines,  including standard fully connected artificial neural networks (ANN), recurrent neural networks (RNN) with the LSTM cell, and recurrent graph neural networks (RGrN)~\cite{jia2021physics}. All of  these baselines do not consider the reservoir release information. 

\textit{Q2:  Can we improve the predictive performance by leveraging the release data from some reservoirs?} In $S_1$, we compare multiple variants of the proposed method, including SAG with the PP approach for both Cannonsville and Pepacton Reservoirs (SAG-pp), SAG with the PP approach for Cannonsville and SE approach for Pepacton (SAG-ppC, assuming  access to the release information for Pepacton), SAG with the PP approach for Pepacton and SE approach for Cannonsville (SAG-ppP, assuming  access to the release information for Cannonsville), 
and SAG with the SE approach for both reservoirs (SAG-sim). We also implement another version of  SAG with the SE approach for both reservoirs (SAG-flow), which uses the release flow information in the SE approach but does not use physical simulations $\textbf{u}$ (Eq.~\ref{eq:se}).  

\textit{Q3:  As an extension to Q2, can we leverage the stream-reservoir patterns learned from a well-monitored region (where we have release information) to improve the prediction for a spatially disjointed region where we do not have reservoir release information?}  In this test, we implement a single global model SAG-ppN which uses the SE approach for two reservoirs in $S_1$ (assuming we have the release data from $S_1$) and the PP approach for the Neversink Reservoir in $S_2$. We compare it with ANN, RNN, and RGrN. We also test the SAG-flow model, which uses the release information of both $S_1$ and $S_2$ through the SE approach. 
For all the tests, we use data from Jan 01, 1980, to Dec 25, 2006, for training and then measure the testing performance on data from Dec 26, 2006, to Jun 22, 2020. Here the training and testing periods are selected to maintain a 2:1 ratio over all the time steps.




\subsection{Predictive performance without using release data}

In Table~\ref{tab:test1}, we summarize the performance of different methods in terms of their overall RMSE, and the RMSE on two segments Seg$_C$ and Seg$_P$. These two segments are selected based on three criteria: (1) Seg$_C$ and Seg$_P$ are downstream from Cannonsville and Pepacton Reservoirs, respectively, (2) they are closest to reservoirs in terms of stream distance, and (3) they have more than 500 observations for evaluation.  We run each test five times with random initialization and report the mean value and the standard deviation of the RMSE.

Table~\ref{tab:test1} shows that SAG-pp performs much better than other baselines. We show the predictions made by RNN, RGrN, and SAG-pp on Seg$_C$ in Fig.~\ref{fig:prd} (a). It can be seen that all the methods over-predict the water temperature at the beginning of the summer period when there is a sudden increase of water release from Cannonsville (Fig.~\ref{fig:prd} (b)). 
However, the water temperature predicted by SAG-pp quickly recovers and is more consistent with observed water temperatures for the remaining summer period when the water release from the reservoir is steady.

\begin{figure*} [!t]
\centering
\includegraphics[width=1.8\columnwidth]{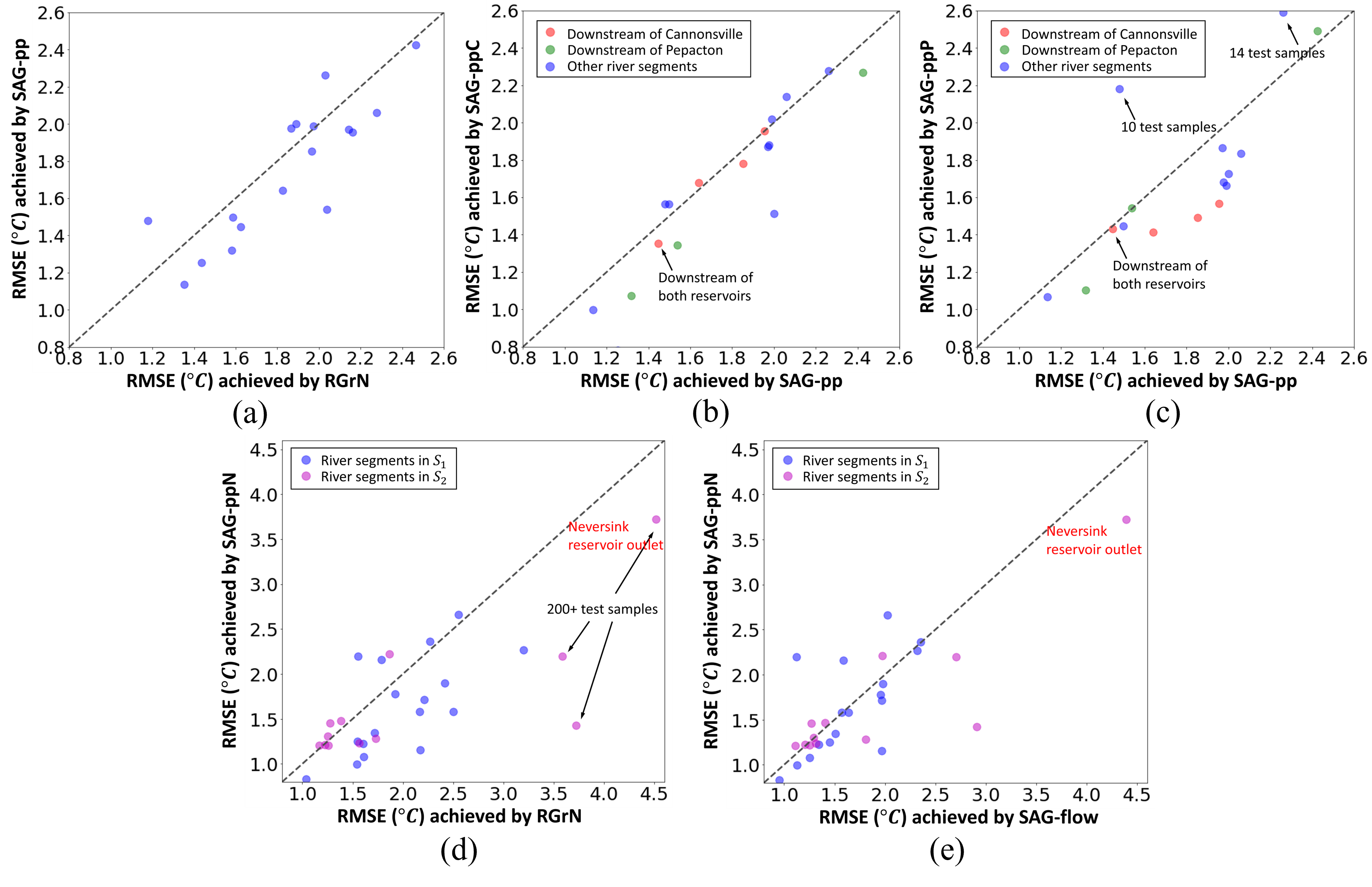}
\vspace{-.2in}
\caption{The RMSE comparison between a pair of methods on each river segment. Here we only show the river segments with more than five testing observations. }
\label{fig:scatter}
\vspace{-.2in}
\end{figure*}

Both SAG-pp and RGrN do not use the release information, and they both consider the connections amongst all the river segments. The better performance of SAG-pp over RGrN confirms the effectiveness of the PP embeddings.  Fig.~\ref{fig:scatter} (a) also shows that SAG-pp  performs better than RGrN for most river segments.

\begin{table}[!t]
\small
\newcommand{\tabincell}[2]{\begin{tabular}{@{}#1@{}}#2\end{tabular}}
\centering
\caption{Prediction RMSE ($\pm$ standard deviation) in $S_1$ assuming no release data is available.}
\begin{tabular}{l|cccc}
\hline
\textbf{Method} & Overall & Seg$_C$ & Seg$_P$  \\ \hline 
ANN & 2.04$\pm$0.04  & 2.31$\pm$0.06 & 1.97$\pm$0.04\\
RNN & 1.91$\pm$0.04  &  2.20$\pm$0.06  &  1.71$\pm$0.05 \\
RGrN &1.84$\pm$0.04 &2.08$\pm$0.05 & 1.86$\pm$0.07   \\ \hline
SAG-pp & 1.63$\pm$0.03 & 1.96$\pm$0.05  & 1.56$\pm$0.05 \\
\hline
\end{tabular}
\label{tab:test1}
\end{table}

\subsection{Leveraging available flow data and physical simulations}

In this analysis, SAG-ppC and SAG-ppP achieve better performance than SAG-pp (Table~\ref{tab:test2}).  In Fig.~\ref{fig:scatter} (b) and (c), we show the per-segment comparison between SAG-pp and SAG-ppC, and between SAG-pp and SAG-ppP, respectively. We also mark the segments in different colors if they are downstream from Cannonsville or Pepacton Reservoirs. We can observe that the improvement of SAG-ppC or SAG-ppP over SAG-pp is especially obvious for segments downstream from the ``transparent'' reservoir. For example, after we use the release information for Cannonsville (through the SE approach), the SAG model (SAG-ppP) achieves much better performance on segments downstream from Cannosville (e.g., Seg$_C$) compared to SAG-pp (not using any release data). Moreover, the use of release flow and simulations slightly improves the performance for other segments (not downstream from the ``transparent'' reservoir), which confirms the effective transfer of useful knowledge from the ``transparent'' reservoir to other regions. Although SAG-ppP performs much worse than SAG-pp on two stream segments (Fig.~\ref{fig:scatter} (c)), they both have  few testing samples, which makes the assessment on them less stable.

Additionally, we see  that SAG-flow and SAG-sim performs better than other methods, which meets our expectation since they directly use the release data for both reservoirs. We also show the predictions of \{SAG-pp, SAG-flow, SAG-sim\} on Seg$_C$ and Seg$_P$ in Figs.~\ref{fig:prd} (c) and (d), respectively. We can see that SAG-flow and SAG-sim can better predict the temperature even under a sudden increase of reservoir release.  The improvement from SAG-flow to SAG-sim confirms the effectiveness of incorporating physical simulations in the SE process. 

\begin{table}[!t]
\small
\newcommand{\tabincell}[2]{\begin{tabular}{@{}#1@{}}#2\end{tabular}}
\centering
\caption{Prediction RMSE ($\pm$ standard deviation) after using  the reservoir release information in $S_1$.}
\begin{tabular}{l|cccc}
\hline
\textbf{Method} & Overall & Seg$_C$ & Seg$_P$  \\ \hline 
SAG-pp & 1.63$\pm$0.03 & 1.96$\pm$0.05  & 1.56$\pm$0.05 \\
SAG-ppC & 1.54$\pm$0.04 & 1.91$\pm$0.06  & 1.35$\pm$0.05 \\
SAG-ppP &1.44$\pm$0.04 & 1.57$\pm$0.05 &  1.50$\pm$0.05   \\ 
SAG-flow& 1.43$\pm$0.03 &1.59$\pm$0.04 &1.42$\pm$0.04 \\
SAG-sim& 1.31$\pm$0.02 &1.31$\pm$0.05  &1.29$\pm$0.04\\
\hline
\end{tabular}
\label{tab:test2}
\end{table}

\begin{table}[!t]
\small
\newcommand{\tabincell}[2]{\begin{tabular}{@{}#1@{}}#2\end{tabular}}
\centering
\caption{Prediction RMSE ($\pm$ standard deviation) in $S_1$ and $S_2$.}
\begin{tabular}{l|cccc}
\hline
\textbf{Method} & $S_1$ & $S_2$  \\ \hline 
ANN &  2.12$\pm$0.04 & 1.63$\pm$0.04\\
RNN &  1.89$\pm$0.04 &1.61$\pm$0.04\\
RGrN &  1.88$\pm$0.03 & 1.60$\pm$0.03 \\ \hline
SAG-ppN &  1.37$\pm$0.02 & 1.49$\pm$0.04 \\
SAG-flow & 1.45$\pm$0.03 & 1.53$\pm$0.03 \\
\hline
\end{tabular}
\label{tab:test3}
\end{table}

\subsection{Generalization over space}
SAG-ppN performs better than ANN, RNN, and RGrN in generalizing to a new region (Table~\ref{tab:test3}). In Fig.~\ref{fig:scatter} (c) and (d), we also show the per-segment comparison between RGrN and SAG-ppN, and between SAG-flow and SAG-ppN, respectively.  It is interesting to see that SAG-ppN also outperforms SAG-flow for both $S_1$ and $S_2$ since SAG-flow uses the release information for both $S_1$ and $S_2$ while SAG-ppN only uses the release information in $S_1$. This suggests that (1) the use of release information of $S_1$ can help the model to learn a better representation of PP embedding in $S_2$ (SAG-ppN outperforms SAG-flow on $S_2$), and (2) the use of pseudo-prospective embedding of $S_2$ can also help the model to learn a better representation of reservoir release in $S_1$ (SAG-ppN outperforms SAG-flow on $S_1$).

\section{Acknowledgements}
Xiaowei Jia and Shengyu Chen were supported by a Cooperative Agreement with U.S. Geological Survey via the Great Lakes Cooperative Ecosystem Studies Unit, CESU Grant G21AC10207. Samantha Oliver and Alison Appling were supported by the National Water Quality Program in the U.S. Geological Survey. We thank Tadgh Moore for sharing his experience and insights on process-based reservoir modeling and Hayley Corson-Dosch and Jordan Read for substantial contributions to preparation of those models. 
Any use of trade, firm, or product names is for descriptive purposes only and does not imply endorsement by the U.S. Government.

\section{Conclusion}

In this paper, we build a new method for predicting water temperature of all the river segments in stream networks with reservoirs. We start with creating an SAG model to represent the evolution of streams and reservoirs as well as their interactions.  
We also build a parallel embedding structure for representing the impact of reservoirs, i.e., the model uses the PP approach when the reservoir release data is not available, and uses the SE approach otherwise. The SE approach also leverages the physical knowledge encoded by the physics-based model in embedding the impact of reservoirs. Such knowledge is transferred to other reservoirs through the parallel embedding structure. Our method has been shown to produce improved performance compared to existing methods for three scenarios: (1) when release data are missing for all the reservoir, (2) when the release data are missing for a subset of reservoirs, and (3) when the model is applied to a disjointed stream network with reservoirs (no release data available).

While our method performs  better than existing models, it remains limited if  water managers use different water release criteria for different reservoirs or for different time periods. Moreover, our PP method only considers the anticipated temperature on the next day while the water managers for certain reservoirs may refer to multiple future days for determining the release data. 
Investigation into alternate methods of modeling for various water release criteria and a range of time periods in reservoirs could better inform downstream temperature responses.

\bibliographystyle{unsrt}
\bibliography{reference}

\end{document}